\pdfoutput=1
\documentclass[11pt]{article}

\usepackage[preprint]{acl}

\usepackage{times}
\usepackage{latexsym}

\usepackage[T1]{fontenc}
\usepackage[utf8]{inputenc}

\usepackage{microtype}

\usepackage{inconsolata}

\usepackage{graphicx}

\usepackage{svg}
\usepackage{amsmath}
\usepackage{algorithm}
\usepackage{algorithmic}
\usepackage{amssymb}
\usepackage{adjustbox}
\usepackage{booktabs}
\usepackage{multirow}
\usepackage[table,xcdraw]{colortbl}
\usepackage{colortbl}
\usepackage{enumitem}
\setlist[itemize]{labelindent=0pt,leftmargin=*,itemsep=0pt,topsep=0pt}
\usepackage{fontawesome}
\title{ToxiGAN: Toxic Data Augmentation \\via LLM-Guided Directional Adversarial Generation}

\author{Peiran Li$^{1,2}$, Jan Fillies$^{1,2,3,4}$, Adrian Paschke$^{1,2,3}$\\
    $^1$Freie Universität Berlin, Berlin, Germany \\
    $^2$Fraunhofer-Institut für Offene Kommunikationssysteme, Berlin, Germany \\
    $^3$Institut für Angewandte Informatik, Leipzig, Germany \\
    $^4$Stanford University, Stanford, USA \\
    \texttt{peiran.li@fu-berlin.de}\\}

\begin{document}
\maketitle
\begin{abstract}
Augmenting toxic language data in a controllable and class-specific manner is crucial for improving robustness in toxicity classification, yet remains challenging due to limited supervision and distributional skew. We propose ToxiGAN, a class-aware text augmentation framework that combines adversarial generation with semantic guidance from large language models (LLMs). To address common issues in GAN-based augmentation such as mode collapse and semantic drift, ToxiGAN introduces a two-step directional training strategy and leverages LLM-generated neutral texts as semantic ballast. Unlike prior work that treats LLMs as static generators, our approach dynamically selects neutral exemplars to provide balanced guidance. Toxic samples are explicitly optimized to diverge from these exemplars, reinforcing class-specific contrastive signals. Experiments on four hate speech benchmarks show that ToxiGAN achieves the strongest average performance in both macro-F1 and hate-F1, consistently outperforming traditional and LLM-based augmentation methods. Ablation and sensitivity analyses further confirm the benefits of semantic ballast and directional training in enhancing classifier robustness.
\end{abstract}
%%%%%%%%%%%%%%%%%%%%%%%%%%%%%%%%%%%%%%%%%%%%%%%%%%%%%%%%%%%%%%%%%%%%%%%%%%%%%%%%%
\section{Introduction}
\label{sec:intro}

From comment sections on social media to online gaming chats, toxic language remains alarmingly pervasive, often escaping automated moderation systems. The propagation of such content poses a critical challenge for content moderation, societal safety, and responsible AI development~\cite{wilson2020hate}. Automatic detection systems have shown promise in addressing this issue, yet their performance is often hindered by distributional imbalance in training data, most notably, an overrepresentation of neutral or non-toxic samples~\cite{isaksen2020using, zampieri-etal-2019-semeval, davidson2017automated}. This imbalance can lead to majority-class overfitting and poor generalization, especially in low-resource or emerging domains. As a remedy, data augmentation using synthetically generated toxic examples has gained traction for balancing datasets and improving classifier robustness~\cite{rizos2019augment}.\par
Yet, turning this solution into practice is far from trivial. Generating toxic text for augmentation is a sensitive and technically challenging task~\cite{vidgen2019challenges}. Synthetic examples must be toxic enough to reflect their target label, while maintaining semantic coherence and linguistic realism to ensure training utility~\cite{rizos2019augment}. Uncontrolled generation, particularly from language models or GANs, often leads to samples that are keyword-toxic but semantically incoherent or stylistically inconsistent~\cite{gehman2020realtoxicityprompts}. Moreover, traditional GAN-based approaches suffer from mode collapse and semantic drift~\cite{yu2017seqgan, caccia2018language}, which further compromise sub-mode coverage and authenticity of the generated data, ultimately weakening decision-boundary calibration across toxic subtypes.\par
Given their remarkable fluency and contextual capabilities~\cite{brown2020language}, large language models (LLMs) may appear well-suited for toxic text augmentation. However, their application is usually constrained by safety alignment objectives~\cite{ouyang2022training}. LLMs are designed to resist producing toxic outputs, and when prompted to do so, tend to yield overly sanitized or generic responses~\cite{achiam2023gpt, touvron2023llama}. Consequently, they are limited in their ability to serve as direct toxic text generators, especially for class-specific data augmentation.\par
To address the limitations of existing generation methods, we introduce \textbf{ToxiGAN}\footnote{\scriptsize\url{https://github.com/Peiran-Li-DS/ToxiGAN}}, a controllable toxic text augmentation framework. Rather than using LLMs to generate toxic content directly, ToxiGAN leverages LLM-generated neutral exemplars as \textit{semantic ballast} (reference anchors in embedding space). The generator learns to increase toxicity by deviating from these neutral anchors, while a class-aware discriminator enforces alignment with the target label. This design steers generation toward toxic content that preserves sub-mode coverage and remains label-consistent, supporting decision-boundary calibration across toxic subtypes. To mitigate semantic drift and mode collapse, where outputs either gravitate toward neutral semantics or collapse to a narrow toxic niche, we apply a two-step alternating optimization strategy that separately updates semantic deviation and class discrimination. Our main contributions are as follows:

\begin{itemize}
\item We propose \textbf{ToxiGAN}, a controllable augmentation framework that uses LLM-generated neutral text as \textit{semantic ballast}, to guide stable and diverse generation.
\item We design a two-step alternating directional learning algorithm that separates semantic deviation from class alignment, improving training stability and control.
\item We evaluate ToxiGAN on four hate classification benchmarks and show that it achieves the best average macro-F1 and hate-F1 among GAN- and LLM-based augmentation methods.
\end{itemize}

These results highlight the value of class-aware adversarial generation guided by neutral semantic anchors, enabling effective and scalable toxic data augmentation for real-world classification tasks.

%%%%%%%%%%%%%%%%%%%%%%%%%%%%%%%%%%%%%%%%%%%%%%%%%%%%%%%%%%%%%%%%%%%%%%%%%
\section{Related Work}

\paragraph{Conventional Toxic Text Generation.} % Related to GAN
Generating toxic or hateful text in a controlled manner has been explored through both supervised and adversarial paradigms. Early work relies on supervised models with prompting or conditional decoding~\cite{gehman2020realtoxicityprompts,sheng2019woman}, but these often lack diversity and controllability. Adversarial frameworks, particularly GAN-based approaches, have emerged as alternatives for generating class-conditioned toxic text. SentiGAN~\cite{wang2018sentigan} introduces a sentiment-controlled generator-discriminator architecture, but struggles to produce coherent long-form samples for abstract or domain-specific categories, especially when distributional gaps across targets are large in the same sentiment. CatGAN~\cite{liu2020catgan} extends the GAN framework to multi-category text generation by introducing a category-aware discriminator and hierarchical training strategy. This allows the model to better handle diverse category labels compared to SentiGAN. However, it remains limited in its reliance on discriminator-only feedback, without leveraging external semantic guidance. Moreover, it is time-consuming and requires complicated tuning in its evolutionary training~\cite{li2023feature}. HateGAN~\cite{cao2020hategan} further adapts adversarial training to hateful text synthesis, focusing on stylistic features and linguistic variation. But it is constrained to binary classification setups (toxic vs. non-toxic) and lacks scalability to multi-class toxicity generation tasks. Our framework builds upon these insights by integrating LLMs as guidance modules and using a two-step alternating directional learning approach to maintain class consistency and generation quality.

\paragraph{LLMs in Text Augmentation.} % Related to LLM guidance
Recent advances in LLMs have enabled them to serve as effective tools for data augmentation~\cite{ye2022zerogen,yoo2021gpt3mix}, especially in low-resource or few-shot settings. While many prior works use LLMs as standalone generators or annotators~\cite{min2022rethinking,sanh2021multitask}, few explicitly integrate them into structured adversarial training pipelines. More importantly, the major deployment-ready LLMs incorporate strict content moderation and safety alignment~\cite{ouyang2022training, ganguli2022red}, which significantly limits their ability to generate or simulate toxic language, even when intended for research or augmentation. Although recent efforts such as ToxiCraft~\cite{hui2024toxicraft} and ToxiLab~\cite{hui2024toxilab} attempt to generate toxic language directly from LLMs using prompt engineering or auxiliary control modules, these methods are often unstable and limited by content filtering policies and lack robustness in generating diverse, controllable toxic samples. In contrast, our work leverages LLMs not only to generate high-quality neutral examples but also to guide semantic direction and assist in discriminator training. This LLM-as-ballast design improves stability and semantic control during adversarial optimization.

\paragraph{Mode Collapse and Semantic Drift in Text Generation.} % Related to Alternating Directional Learning
GAN-based text generation often suffers from mode collapse and semantic drift~\cite{che2017maximum, goodfellow2014generative, spataru2024know}, which erode sub-mode coverage and class fidelity, especially in tasks involving toxic language. Prior work addresses these issues through diversity-promoting objectives~\cite{zhu2018texygen} or classifier-based rewards~\cite{yu2017seqgan}, yet such methods lack semantic grounding. We introduce a semantic anchor via LLM-generated neutral exemplars and mitigate both collapse (concentration into a narrow toxic niche) and drift (gravitating toward neutral semantics) through alternating directional optimization that decouples semantic deviation from class discrimination~\cite{zhang2019addressing, dathathri2019plug, spataru2024know}.

%%%%%%%%%%%%%%%%%%%%%%%%%%%%%%%%%%%%%%%%%%%%%%%%%%%%%%%%%%%%%%%%%%%%%%%%%
\section{Methodology}

Building on the high-level overview in Section~\ref{sec:intro}, we present the full formulation of \textbf{ToxiGAN}, including its architectural components and training dynamics.

\subsection{Problem Formulation}

Let $\mathcal{D}_{real} = \{(x_i, y_i)\}$ denote the training dataset, where $x_i$ is a text input and $y_i \in \{\text{neutral}, \text{toxic}_1, \dots, \text{toxic}_K\}$. Our goal is to train a generator module $G$ that, given a class label $y_k$ and random noise $z \sim P_z$, generate a toxic sample that is (1) semantically authentic, and (2) representative of toxic class $y_k$.

\subsection{Overall Framework}

\begin{figure}[b]
\centering
\includegraphics[width=0.43\textwidth]{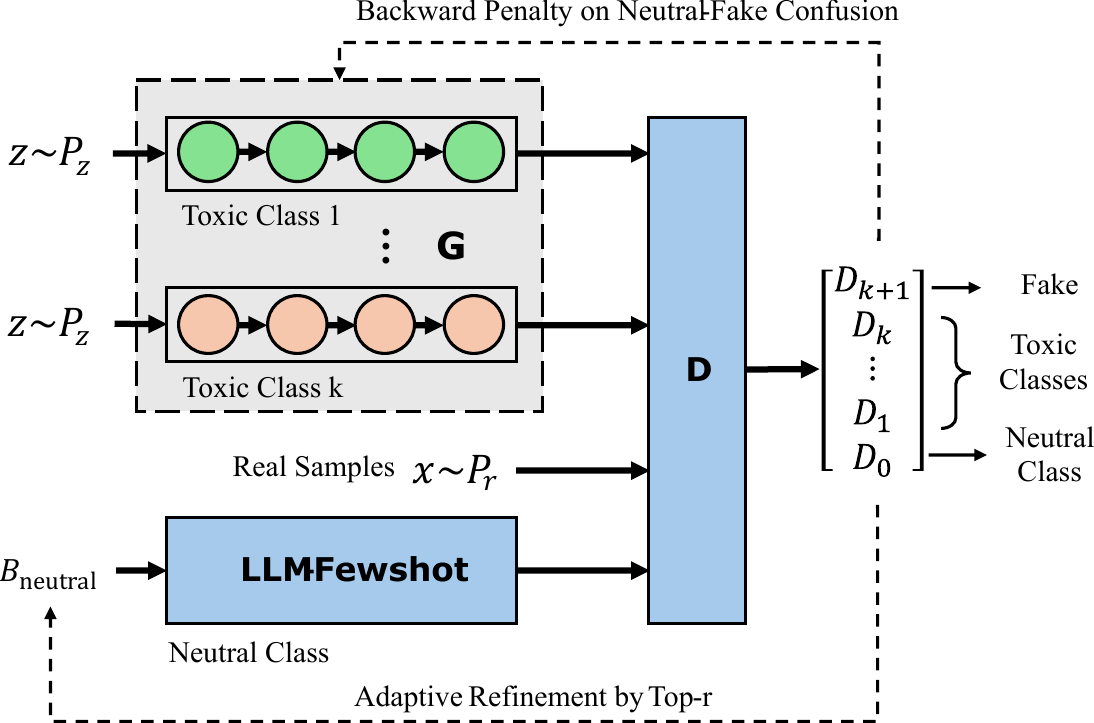} % Reduce the figure size so that it is slightly narrower than the column.
\caption{ToxiGAN with k toxic generators, one neutral texts provider, and one multi-class discriminator.}
\label{fig:framework}
\end{figure}

Figure~\ref{fig:framework} illustrates the overall architecture. ToxiGAN consists of the following components:

\begin{itemize}
    \item \textbf{Toxic Generator Module ($G$)}: Consists of multiple LSTM-based toxic generators and learns to generate samples for each toxic class from a noise distribution. Each class has a dedicated decoding branch.
    \item \textbf{Multi-class Discriminator ($D$)}: Classifies input text into $K+2$ classes: $K$ toxic classes, one neutral class, and one fake class to capture unrealistic generations.
    \item \textbf{LLM-based Neutral Text Provider}: A pre-trained LLM (e.g., Llama 3.2) is used to generate neutral in-domain examples for training $D$ and guiding $G$ via few-shot learning from the real neutral texts.
\end{itemize}

Real and generated samples are passed through $D$ during training. A backward penalty is applied on the ``fake'' and ``neutral'' evaluations to encourage clearer decision boundaries and better generation quality.

\subsection{LLM as Ballast: Preventing Mode Collapse and Semantic Drift}

To address mode collapse and semantic drift, we introduce a \textbf{LLM-based neutral text provider} that offers high-quality, fluent in-domain exemplars. These samples serve as \textit{semantic anchors} during both generation and discrimination, improving stability and realism under domain shifts while \textbf{preserving sub-mode coverage} in representation space.\par
It contributes in three complementary ways: (1) \textbf{Neutral Text Generation}: using a small set of real examples as prompts, the LLM generates fluent, contextually appropriate neutral samples. These exemplars act as semantic ballast for both the generator and discriminator. (2) \textbf{Discriminator Enhancement}: LLM-generated neutral samples are included during discriminator training to sharpen its separation of target classes and authenticity, which supports more reliable decision-boundary calibration in low-resource or noisy regimes. (3) \textbf{Semantic Filtering}: when the neutral data are noisy or partially mislabeled, the LLM provides a soft constraint that downweights samples inconsistent with natural-language regularities, mitigating drift and maintaining coverage across toxic sub-modes. Taken together, these roles help prevent collapse and drift, preserve sub-mode coverage and authenticity, and yield label-faithful toxic text that better supports downstream boundary calibration.

\paragraph{Adaptive Refinement of Neutral Pool.}
\label{par:ballast}
To ensure semantic divergence is measured against high-quality anchors, we employ a dynamic filtering strategy guided by discriminator evaluation of neutral class $D_0$. Starting from a large pool $\mathcal{X}_{\text{neutral}}$ of real neutral texts, we compute per-sample neutrality scores: $s(x) = D_0(x)$,
where $D_0(x)$ reflects how similar the perceived $x$ is to the real neutral data evaluated by the discriminator. After each adversarial epoch, we retain only the top-$r\%$ of neutral candidates (by $s(x)$), halving $r$ until a fixed-size ballast pool (e.g., 100 samples) is reached:

\begin{equation}
\label{eq:shotSelect}
\scalebox{0.9}{$
\mathcal{B}_{\text{neutral}}^{(t)} = \text{Top}_r \left( \mathcal{X}_{\text{neutral}}, s(x) \right)
$}
\end{equation}
All LLM few-shot prompts are drawn from this final refined pool.

\subsection{Two-Step Alternating Directional Learning}

While prior work e.g. SentiGAN has explored using penalty-based function, it simply relies on evaluating how well the synthetic text aligns with in-domain authenticity by the discriminator. In the context of toxic text generation, it is not sufficient to merely generate text classified as ``authentic'' by a discriminator; the generated output should also semantically diverge from neutral language in a meaningful and controlled direction.

\begin{figure}[]
\centering
\includegraphics[width=0.35\textwidth]{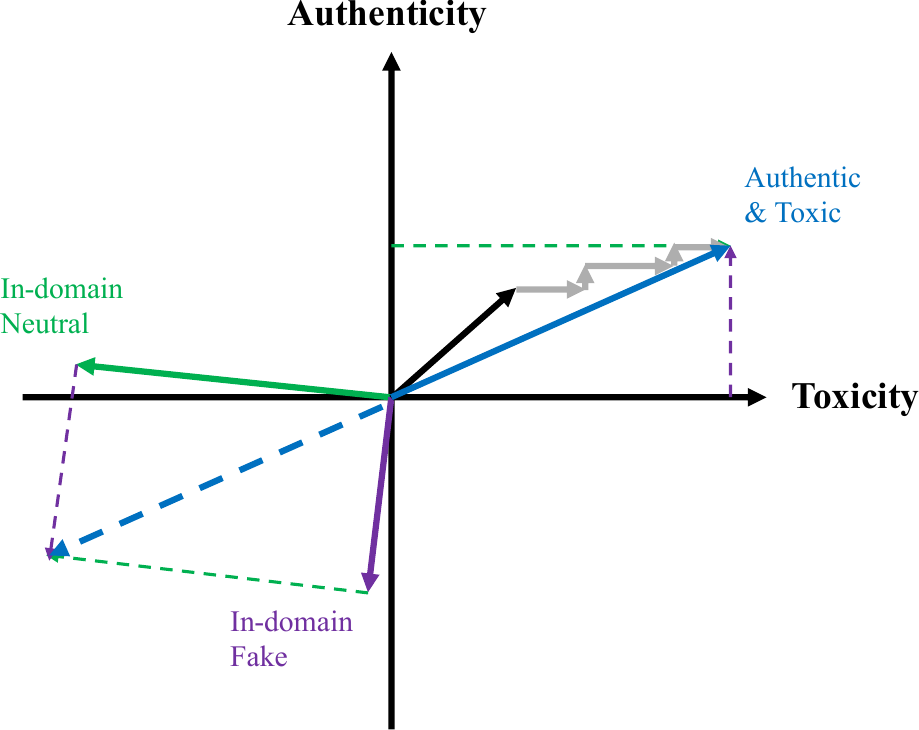} % Reduce the figure size so that it is slightly narrower than the column.
\caption{Illustration of Two-Step Alternating Directional Learning in Embedding Space. The black arrow shows the initial generation after pretraining. Gray arrows represent updates during alternating optimization: shifting toward toxicity and authenticity directions by penalizing unexpected directional evaluations.}
\label{fig:2d}
\end{figure}

One of our core contributions is a semantic directional constraint that guides generation in embedding space, as shown in Figure~\ref{fig:2d}. Instead of rewarding toxic and authentic outputs jointly, we propose a Two-Step Alternating Directional Learning strategy that disentangles and alternates two core training objectives: semantic toxicity and linguistic authenticity. Crucially, these objectives rely on distinct evaluation signals, cosine distance from neutral exemplars and class probabilities from the discriminator, making joint optimization nontrivial. Alternating updates allow each direction to be optimized according to its own metric, preserving interpretability while promoting both control and domain-authenticity.\par

\paragraph{Alternating vs.\ joint objective.}
A natural alternative to our alternating toxicity and authenticity steps is to optimize a single
joint objective for the generator:
\begin{equation}
  \mathcal{L}^{\text{joint}}_{G_i}
    = \lambda \,\mathcal{L}^{\text{tox}}_{G_i}
    + (1 - \lambda)\,\mathcal{L}^{\text{auth}}_{G_i},
  \label{eq:joint-loss}
\end{equation}
where $\mathcal{L}^{\text{tox}}_{G_i}$ corresponds to the toxicity-driven divergence objective (Eq.~\ref{eq:ToxStep}), $\mathcal{L}^{\text{auth}}_{G_i}$ corresponds to the authenticity objective (Eq.~\ref{eq:DisStep}), and $\lambda \in [0,1]$ controls their relative importance. However, as illustrated by the loss dynamics in Figure~\ref{fig:loss_trends} (Appendix \ref{Optimization}), the scale and evolution of $\mathcal{L}^{\text{tox}}_{G_i}$ and $\mathcal{L}^{\text{auth}}_{G_i}$ are not well aligned during training: at different stages, one term can dominate the other both in magnitude and in gradient variability. In such a setting, a fixed global weighting $\lambda$ in Eq.~\ref{eq:joint-loss} is difficult to tune and can easily bias the generator toward either overly aggressive toxicity (ignoring authenticity) or overly conservative changes (staying too close to neutral exemplars). We therefore adopt an alternating optimization scheme, where we update $G_i$ with $\mathcal{L}^{\text{tox}}_{G_i}$ in the toxicity step and with $\mathcal{L}^{\text{auth}}_{G_i}$ in the authenticity step, providing a simple and robust way to balance the two objectives over the course of training without committing to a single hand-tuned trade-off parameter.

Let $G_i(z)$ be a generated sample for toxic class $i$ by random noise $z$ and $x_{\text{neutral}}$ be a neutral sentence sampled via LLM few-shot prompting. At each training step $t$, the generator is updated based on one of two alternating objectives:
\begin{itemize}
    \item \textbf{Toxicity Step (odd $t$):} The generator is guided to semantically diverge from neutral content by minimizing the cosine similarity between the generated sentence and a neutral reference:
    \begin{equation}
    \scalebox{0.88}{$
    \label{eq:ToxStep}
    \scalebox{0.85}{$
    \mathcal{L}^{(t)}_{G_i} = \mathbb{E}[\max_{x \in \mathcal{B}_{\text{neutral}}} \cos\left( \Phi(G_i(z)), \Phi(x) \right)], \quad \text{if } t \bmod 2 = 1
    $}
    $}
    \end{equation}
    where $\Phi(\cdot)$ denotes the sentence embedding function (e.g., all-MiniLM-L6-v2). $\mathcal{B}_{\text{neutral}}$  is a set of fluent, LLM-generated neutral sentences serving as \textit{Semantic Ballast}.

    \item \textbf{Authenticity Step (even $t$):} The generator is optimized to improve naturalness and realism by maximizing the discriminator’s belief that the output is real:
    \begin{equation}
    \label{eq:DisStep}
    \scalebox{0.85}{$
    \mathcal{L}^{(t)}_{G_i} = \mathbb{E}[1 - D_i(G_i(z))], \quad \text{if } t \bmod 2 = 0
    $}
    \end{equation}
    where $D_i(\cdot)$ is the discriminator output of toxic class $i$.
\end{itemize}

This alternation allows the generator to progressively move away from LLM-neutral semantics while remaining within the bounds of in-domain authenticity.

%%%%%%%%%%%%%%%%%%%%%%

\subsection{Adversarial Training}

ToxiGAN employs a class-conditional adversarial training framework comprising $K$ class-specific generators $\{G_i\}_{i=1}^{K}$ and a unified multi-head discriminator $D$. The training process alternates between two optimization goals: promoting semantic divergence from neutral anchors (toxicity direction) and aligning with real data distribution (authenticity), as illustrated in Algorithm~\ref{alg:adv-training} (Appendix \ref{algo}).

Each generator $G_i$ is initialized via maximum likelihood estimation (MLE) pretraining on real toxic samples from class $i$. In parallel, the discriminator $D$ is pre-trained using a mixture of real, generated, and LLM-synthesized neutral texts. Neutral exemplars are dynamically selected from a refined ballast pool $\mathcal{B}^{(t)}_{\text{neutral}}$ (as previously described), ensuring adaptive semantic contrast throughout training. This design maintains meaningful toxicity direction signals and improves class fidelity over static or randomly sampled baselines.

During adversarial training, at each epoch $t$, we iterate over classes $i \in \{1, \dots, K\}$ and generate toxic texts $\mathcal{F}_i$ from $G_i$. The generator objective alternates across epochs:
\begin{equation}
\label{eq:obj_g}
\scalebox{0.83}{$
J_{G_i}(\theta_{g_i}) = \mathcal{L}^{(t)}\big(G_i(z; \theta_{g_i})\big)
$}
\end{equation}
where $\mathcal{L}^{(t)}$ corresponds to a toxicity-inducing loss at odd steps (see Eq.~\ref{eq:ToxStep}) and an authenticity-promoting loss at even steps (see Eq.~\ref{eq:DisStep}). This two-step directional optimization prevents semantic drift and encourages category fidelity across generation stages.

The discriminator $D$ receives three types of inputs: (1) real labeled data from each class $i$, (2) synthetic toxic texts from $\{G_i\}$, and (3) neutral texts generated by the $\mathcal{LLM}$. It consists of $K+2$ output heads: one per toxic class ($D_i$), one for fake samples ($D_{k+1}$), and one for LLM-neutral detection. Its training objective is:
\begin{equation}
\label{eq:dis}
\scalebox{0.83}{$
\begin{aligned}
J_D(\theta_d) =\ & - \mathbb{E}_{z \sim P_z} \big[\log D_{k+1}(G_i(z); \theta_d)\big] \\
& - \mathbb{E}_{\tilde{x} \sim \mathcal{LLM}(\mathcal{B}_{\text{neutral}}^{(t)})} \big[\log D_{k+1}(\tilde{x}; \theta_d)\big] \\
& - \sum_{i=1}^K \mathbb{E}_{x \sim P_{r_i}} \big[\log D_i(x; \theta_d)\big]
\end{aligned}
$}
\end{equation}
After each epoch, the ballast pool $\mathcal{B}^{(t)}_{\text{neutral}}$ is dynamically refined by filtering candidate prompts based on discriminator confidence scores (Eq (\ref{eq:shotSelect})). This mechanism ensures that the LLM continues to provide diverse and semantically representative neutral anchors, supporting both stable optimization and class-specific control.

\section{Experimentation}

%%%%%%%%%%%%%%%% Dataset Description %%%%%%%%%%%%%%%%%%%%%%%%%
\begin{table}[b]
\adjustbox{scale=0.65}{
\begin{tabular}{cp{0.9cm}lp{5.5cm}}
\toprule
\textbf{Dataset} & \textbf{Source} & \textbf{Guideline} & \textbf{Category} \\
\midrule
WZ & Twitter & Hate Targets & \textbf{3}-class: racism, sexism, neither \\
DC & Discord & Linguistic Forms & \textbf{7}-class: no-hate, stereotype, dehumanization, violence, discrimination, irony, slander \\
HX & Twitter, Gab & Hate Targets & \textbf{5}-class: normal / offensive, general-hate, gender / sex, race, religion \\
OR & Reddit & Initiate or not & \textbf{3}-class: non-hate, initiating, responding\\
\bottomrule
\end{tabular}}
\caption{Overview of the datasets. Categories cover a diverse range of toxicity taxonomies.}
\centering
\label{tab:dataset}
\end{table}

\subsection{Experiment Setup} % Datasets and task settings

\paragraph{Datasets.} We evaluate our approach on four publicly available hate speech datasets with diverse origins and annotation schemes:

\textbf{WZ}~\cite{waseem2016hateful} contains tweets annotated by experts into \textit{racism}, \textit{sexism}, or \textit{neither}, and is widely used for binary and multiclass toxic language detection.

\textbf{DC} (Discord Chat)~\cite{fillies2023hateful} is collected from gaming chat communities and annotated along linguistic dimensions such as \textit{stereotype}, \textit{violence}, \textit{normalized discrimination}, \textit{slander}, and \textit{irony}, providing a fine-grained perspective on in-domain toxicity styles.

\textbf{HX} (HateXplain)~\cite{mathew2021hatexplain} combines social media posts from Twitter and Gab, annotated through crowd-sourced rationales, and categorized into \textit{offensive}, \textit{general hate}, and targeted categories such as \textit{gender/sex}, \textit{race}, and \textit{religion}.

\textbf{OR} (Offensive Reddit)~\cite{qian2019benchmark} introduces a structured Reddit dataset with conversational dynamics. Each instance is labeled as \textit{non-hate}, \textit{initiating-hate}, or \textit{responding-hate}, reflecting intervention scenarios in real-world moderation.

Table \ref{tab:dataset} summarizes the key attributes of these datasets after processing. Together, they span multiple source domains (e.g., Twitter), annotation paradigms, and toxicity taxonomies, allowing us to comprehensively evaluate both the controllability and generalizability of ToxiGAN.

\paragraph{Baselines.}
We compare ToxiGAN against a broad set of data augmentation methods commonly used in text generation and toxic content detection. These include:

\begin{itemize}
\item \textbf{No Augmentation (Base)}: Training directly on the limited toxic dataset without synthetic data.
\item \textbf{Gold Labels (Ideal)}: A hypothetical upper-bound setting where all original toxic samples are preserved across different ratios.
\item \textbf{Conventional Methods:}
\textbf{Oversampling}: Duplicating real toxic samples to match the target augmentation ratio. \textbf{EDA}~\cite{wei2019eda}: A light heuristic method applying synonym replacement and word swapping. \textbf{Back-Translation}: Translating sentences to another language (e.g. German in this case) and back to create paraphrases, via WMT-19 translator\footnote{\scriptsize\url{https://huggingface.co/facebook/wmt19-de-en}}. \textbf{T5-Paraphrase}~\cite{piedboeuf2023chatgpt, scherrer2020tapaco}: Using a fine-tuned T5 model\footnote{\scriptsize\url{https://huggingface.co/hetpandya/t5-small-tapaco}} for paraphrasing toxic data. \textbf{SentiGAN}~\cite{wang2018sentigan}: A GAN-based method that controls sentiment via multiple generators and a discriminator.

\item \textbf{LLM-Based Generation}: Generating toxic samples using \textbf{Mistral-v0.3}\footnote{\scriptsize\url{https://huggingface.co/mistralai/Mistral-7B-Instruct-v0.3}} via \textbf{ZeroGen}~\cite{ye2022zerogen} (ZG; zero-shot synthesis from class definitions and constraints without seed examples), and \textbf{SunGen}~\cite{gao2022self} (SG; ZG followed by self-guided reweighting to downweight noisy synthetic samples), then \textbf{Fewshot} (FS) generation with randomly selected 5 examples in each corresponding toxic class; \textbf{LLaMA3.2}\footnote{\scriptsize\url{https://huggingface.co/meta-llama/Llama-3.2-1B}}, \textbf{GPT-4.1}\footnote{\scriptsize\url{https://platform.openai.com/docs/models/gpt-4.1-nano}}, and \textbf{GPT-4o}\footnote{\scriptsize\url{https://platform.openai.com/docs/models/gpt-4o}} via carefully crafted prompts (\textit{ToxiCraft}~\cite{hui2024toxicraft}, denoted as TC), due to their tight moderation.

\end{itemize}

We split each dataset into 80\% training, 10\% validation, and 10\% testing. To simulate low-resource settings, only 50\% of the training set is used as labeled data; the remaining 50\% is replaced with augmented samples from each method. All results are averaged over 5 runs. %Our ablation study evaluates simplified variants of ToxiGAN by removing either the LLM-based semantic ballast or the directional learning mechanism, in order to assess the individual contribution of each component.

\paragraph{Evaluation Metrics.}
We evaluate model performance using the following metrics:

\begin{itemize}
\item \textbf{Toxicity Score}: The average toxicity scores of a group, computed by external toxicity evaluator\footnote{\scriptsize\url{https://github.com/unitaryai/detoxify}}.
\item \textbf{Macro-F1}: The unweighted average F1-scores across all classes, capturing overall balance.
\item \textbf{Hate-F1}: The F1-score computed specifically for the toxic or hate class, highlighting the model’s ability to detect rare but critical instances.
\end{itemize}

\subsection{Toxicity of Synthetic Texts}
\begin{figure}[]
\centering
\includegraphics[width=0.35\textwidth]{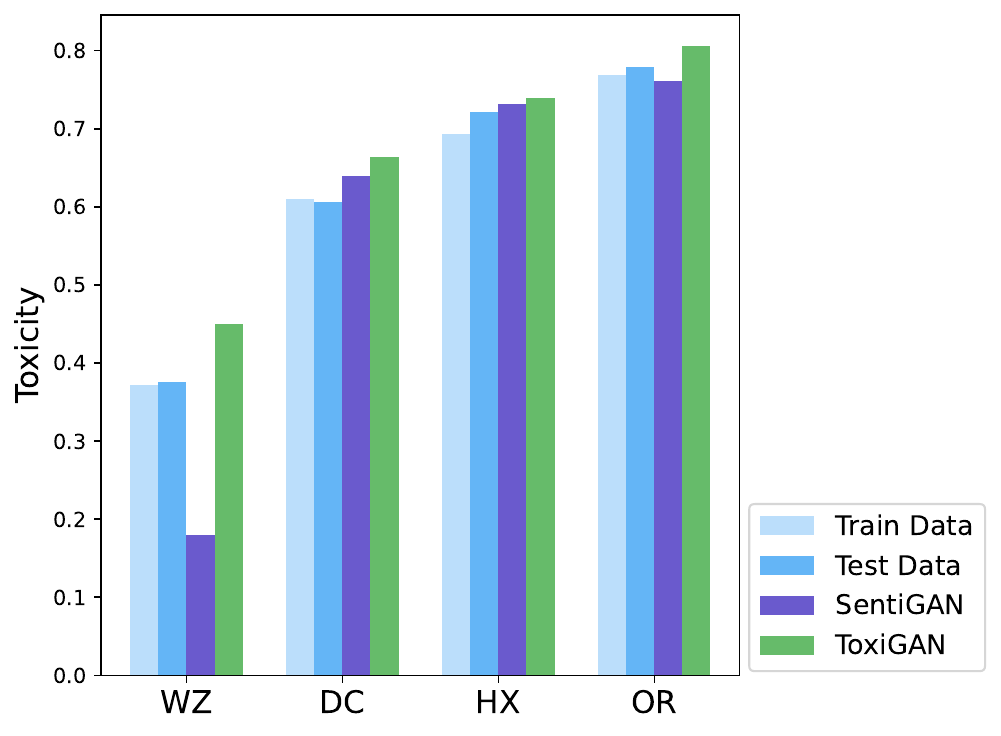} % Reduce the figure size so that it is slightly narrower than the column.
\caption{Average toxicity scores of real and synthetic samples across datasets.}
\label{fig:tox_bar}
\end{figure}

To assess whether the generated texts exhibit sufficient toxicity, we computed toxicity across four datasets: WZ, DC, HX, and OR. Figure~\ref{fig:tox_bar} compares the toxicity levels of training data, test data, and synthetic texts produced by SentiGAN and ToxiGAN.\par
ToxiGAN consistently produces samples with toxicity levels comparable to or higher than those in the original training and testing sets. In contrast, SentiGAN-generated texts often display reduced toxicity, especially in datasets with low toxicity originally, e.g WZ, where SentiGAN has difficulty to ``interpret" how to generate toxic texts. This suggests that ToxiGAN is more effective at preserving the intended toxic signal, contributed by its directional training and LLM-guided neutral anchoring. These results confirm that ToxiGAN not only preserves authenticity but also enhances class-level toxicity control in the generated samples.\par

%%%%%%%%%%%%%%%% Baselines Comparison %%%%%%%%%%%%%%%%%%%%%%%%%
% \definecolor{best}{RGB}{144, 237, 144}   % green
% \definecolor{second}{RGB}{255, 242, 204} % yellow
\definecolor{best}{RGB}{220, 240, 220}   % green
\definecolor{second}{RGB}{245, 240, 220} % yellow

\renewcommand{\arraystretch}{1.25}
\setlength{\tabcolsep}{4pt}

\begin{table*}[t]
\centering
\footnotesize
\adjustbox{max width=0.7\linewidth}{
\begin{tabular}{clccccccccp{0.01cm}cc}
\toprule
& & \multicolumn{2}{c}{\textbf{WZ}} & \multicolumn{2}{c}{\textbf{DC}} & \multicolumn{2}{c}{\textbf{HX}} & \multicolumn{2}{c}{\textbf{OR}} & & \multicolumn{2}{c}{\textbf{\textit{Avg.}}}\\
\cmidrule(lr){3-4} \cmidrule(lr){5-6} \cmidrule(lr){7-8} \cmidrule(lr){9-10} \cmidrule(lr){12-13}
\multirow{-2}{*}{\textbf{Classifier}} & \multirow{-2}{*}{\textbf{Augmentation}} & H.-F1 & M.-F1 & H.-F1 & M.-F1 & H.-F1 & M.-F1 &  H.-F1 & M.-F1 & & H.-F1 & M.-F1  \\
\midrule
\midrule
\multirow{13}{*}{\rotatebox[origin=c]{90}{\textbf{BERT}}}
& Base (No Aug.)          & 68.1 & 74.8 & 26.7 & 34.8 & 34.6 & 41.6 & 46.5 & 61.6 & & 44.0 & 53.2 \\

& Ideal (Gold Labels)       & 69.7 & 76.0 & 28.5 & 35.8 & 37.7 & 43.6 & 49.7 & 63.8 & & 46.4 & 54.8
\\
\cmidrule{2-13}
& Oversampling    & 68.3 & 75.0 & 27.8 & 35.3 & 35.1 & 42.3 & 47.2 & 62.0 & & 44.6 & 53.6 \\
& EDA    & 71.9 & 77.2 & 26.3 & 35.9 & 36.0 & 42.4 & 47.2 & 62.0 & & 45.4 & 54.4 \\
& Back-Translate  & \cellcolor{best}73.4 & \cellcolor{best} 78.4 & 27.0 & 36.4 & 34.7 & 41.8 & 47.5 & \cellcolor{second}62.2 & & \cellcolor{second}45.6 & 54.7 \\
& T5-Paraphase  & 70.4 & 76.3 & 27.1 & 36.8 & 36.0 & 40.7 & \cellcolor{second}47.6 & \cellcolor{second}62.2 & & 45.3 & 54.0 \\
& Mistral-v0.3-ZG & 71.3 & 76.9 & 28.2 & 37.4 & 33.6 & 41.3 & 44.6 & 60.2 & & 44.4 & 54.0  \\
& Mistral-v0.3-SG & 71.9 & 77.2 & \cellcolor{second}28.5 & \cellcolor{second}37.6 & 33.3 & 41.6 & 45.5 & 61.0 & & 44.8 & 54.3 \\
& Mistral-v0.3-FS & 71.5 & 76.8 & 28.0 & \cellcolor{second}37.6 & \cellcolor{second}36.1 & \cellcolor{best}43.1 & 46.6 & 61.5 & & 45.5 & \cellcolor{second}54.8 \\
& Llama3.2-TC     & 71.9 & 77.3 & 27.4 & 37.0 & 34.9 & 40.9 & 47.1 & 61.6 & & 45.3 & 54.2 \\
& GPT4.1-TC       & 71.8 & 77.2 & 26.9 & 35.2 & 26.0 & 37.0 & 46.2 & 60.9 & & 42.7 & 52.6 \\
& GPT4o-TC        & \cellcolor{second}72.6 & 77.6 & 27.5 & 35.7 & 29.3 & 38.4 & 46.2 & 61.1 & & 43.9 & 53.2 \\
%\cmidrule{2-12}
& \textbf{ToxiGAN (Ours)}       & 72.2 & \cellcolor{second}77.7 & \cellcolor{best}29.7 & \cellcolor{best}37.7 & \cellcolor{best}36.9 & \cellcolor{second}42.8 & \cellcolor{best}47.8 & \cellcolor{best}62.7 & & \cellcolor{best}46.7 & \cellcolor{best}55.2 \\

\midrule
\midrule

\multirow{13}{*}{\rotatebox[origin=c]{90}{\textbf{RoBERTa}}}
& Base (No Aug.)   & 71.2 & 76.9 & 29.2 & 38.6 & 39.6 & 44.4 & 45.6 & 61.0 & & 46.4 & 55.2 \\
& Ideal (Gold Label)       & 72.5 & 77.9 & 30.7 & 39.7 & 42.6 & 49.1 & 49.4 & 63.7 & & 48.8 & 57.6 \\
\cmidrule{2-13}
& Oversampling    & 71.7 & 77.3 & \cellcolor{second}29.5 & 38.6 & 39.0 & 45.1 & 46.2 & 61.4 & & 46.6 & 55.6 \\
& EDA    & 71.3 & 77.0 & 29.3 & 39.0 & 40.6 & 47.6 & 46.8 & 62.0 & & 47.0 & 56.4 \\
& Back-Translate  & 72.8 & 77.9 & 29.4 & \cellcolor{second}39.1 & 40.9 & 47.0 & 45.9 & 61.1 & & 47.2 & 56.3 \\
& T5-Paraphase  & \cellcolor{second}73.1 & \cellcolor{best}78.3 & 29.4 & 38.9 & 40.0 & 46.9 & 46.9 & 61.8 & & 47.3 & \cellcolor{second}56.5 \\
& Mistral-v0.3-ZG & 71.3 & 77.0 & 28.2 & 37.6 & 41.9 & 46.8 & 44.9 & 60.6 & & 46.6 & 55.5 \\
& Mistral-v0.3-SG & 72.0 & 77.5 & 29.1 & 38.3 & \cellcolor{second}44.0 & 47.3 & 46.1 & 61.5 & & 47.8 & 56.2 \\
& Mistral-v0.3-FS & 71.7 & 77.2 & 28.2 & 37.9 & \cellcolor{best}44.2 & \cellcolor{best}48.8 & \cellcolor{second}47.5 & \cellcolor{second}62.2 & & \cellcolor{second}47.9 & \cellcolor{second}56.5 \\
& Llama3.2-TC     & 72.8 & 77.7 & 27.5 & 37.3 & 41.0 & 45.7 & 44.5 & 59.7 & & 46.4 & 55.1 \\
& GPT4.1-TC       & 72.4 & 77.6 & 27.3 & 36.3 & 37.4 & 42.6 & 45.5 & 60.6 & & 45.6 & 54.3 \\
& GPT4o-TC        & \cellcolor{best}73.4 & \cellcolor{best}78.3 & 28.7 & 38.1 & 39.1 & 44.4 & 46.6 & 61.5 & & 47.0 & 55.6 \\
%\cmidrule{2-12}
& \textbf{ToxiGAN (Ours)}       & 72.8 & 77.9 & \cellcolor{best}31.0 & \cellcolor{best}40.1 & 41.6 & \cellcolor{second}48.1 & \cellcolor{best}48.3 & \cellcolor{best}62.9 & & \cellcolor{best}48.4 & \cellcolor{best}57.3 \\

\bottomrule
\end{tabular}}
\caption{Augmentation results compared to other baselines. The best and second best are highlighted in green and yellow. Note: LLM-based augmentation methods may be constrained by internal safety alignment, affecting their ability to generate truly toxic samples despite prompt customization.}
\label{tab:augmentation_comparison}
\end{table*}

\subsection{Result of Augmentation Performance}
Table~\ref{tab:augmentation_comparison} reports performance comparisons across four datasets and two backbone classifiers (BERT and RoBERTa), under various data augmentation strategies. We evaluate both Macro-F1 and Hate-F1 to capture class-level balance and minority class performance.\par
\textbf{On average, ToxiGAN outperforms all baselines across both Macro-F1 and Hate-F1}, achieving the best mean results across datasets and classifiers. Notably, on DC and OR datasets with implicit categories (by linguistic forms and initiate or not), ToxiGAN outperforms all baselines in both metrics. This demonstrates its advantage in modeling nuanced toxicity directions. Among traditional augmentation techniques, back-translation and T5-based paraphrasing yield competitive results, while LLM-based generation (e.g., Mistral, GPT-4) shows less consistent improvements.\par
Compared to the Ideal (Gold Label) setting, ToxiGAN closes the performance gap and even surpasses it in many cases. This suggests that semantically guided synthetic samples can be as effective as real annotated data in low-resource scenarios. The improvements are particularly pronounced for Hate-F1, indicating stronger capability in capturing toxic-specific signal.\par

We further observe that several LLM-based augmentation methods (e.g., \textbf{GPT4o}, \textbf{LLaMA3.2}) do not consistently outperform simpler techniques such as back-translation or T5-based paraphrasing. We hypothesize that this may be attributed to the internal moderation mechanisms or alignment procedures embedded in modern LLMs. Despite prompt engineering and the use of frameworks like \textbf{ToxiCraft} to elicit toxic samples, models such as GPT-4 and LLaMA-3 still exhibit reluctance or failure to generate explicitly toxic content. This results in samples that are grammatically fluent but often semantically neutral or diluted, thereby reducing their efficacy in contrastive training. Even models like \textbf{Mistral}, which are not tightly moderated, may still inherit instruction-tuning biases toward politeness or neutrality due to alignment with general-purpose pre-training corpora. These implicit constraints likely inhibit the generation of toxic-specific features, explaining their relatively weaker performance in Hate-F1.

%%%%%%%%%%%%%%%% Ablation Study Table %%%%%%%%%%%%%%%%%%%%%%%%%
\begin{table*}[]
\centering
\adjustbox{max width=0.8\linewidth}{
\begin{tabular}{ccccccccccccccc}
\hline
\toprule
& & & &\multicolumn{2}{c}{\textbf{WZ}} & \multicolumn{2}{c}{\textbf{DC}} & \multicolumn{2}{c}{\textbf{HX}} & \multicolumn{2}{c}{\textbf{OR}} & \multicolumn{2}{c}{\textbf{\textit{Avg.}}}\\
\cmidrule(lr){5-6} \cmidrule(lr){7-8} \cmidrule(lr){9-10} \cmidrule(lr){11-12} \cmidrule(lr){13-14} 
\multirow{-2}{*}{\textbf{Classifier}} & \multirow{-2}{*}{\textbf{Ablation}} & \multirow{-2}{*}{\textbf{Sem. Bal.}}                     & \multirow{-2}{*}{\textbf{Alt.-Dir.}}                      & H.-F1 & M.-F1 & H.-F1 & M.-F1 & H.-F1 & M.-F1 &  H.-F1 & M.-F1 & H.-F1 & M.-F1  \\ \hline
           & w/o LLM ($\Leftrightarrow$ \textit{SentiGAN}) & {\color[HTML]{9A0000} x} &  {\color[HTML]{9A0000} x} & 69.5 & 75.7 & 28.9 & 36.4 & 31.8 & 39.2 & 47.0 & 61.9 & 44.3 & 53.3\\
           & w/o Toxicity Step & {\color[HTML]{009901} \checkmark} & {\color[HTML]{9A0000} x} & 70.9 & 76.6 & 29.6 & 37.1 & 35.0 & 41.8 & 47.4 & 62.2 & 45.7 & 54.4\\
\multirow{-3}{*}{BERT}        & Full ToxiGAN & {\color[HTML]{009901} \checkmark} & {\color[HTML]{009901} \checkmark}  & \textbf{72.2} & \textbf{77.7} & \textbf{29.7} & \textbf{37.7} & \textbf{36.9} & \textbf{42.8} & \textbf{47.8} & \textbf{62.7} & \textbf{46.7} & \textbf{55.2}\\ \hline
           & w/o LLM ($\Leftrightarrow$ \textit{SentiGAN}) & {\color[HTML]{9A0000} x} &  {\color[HTML]{9A0000} x} & 71.6 & 77.0 & 29.4 & 38.4 & 40.5 & 45.9 & 46.6 & 61.9 & 47.0 & 55.8\\
           & w/o Toxicity Step  & {\color[HTML]{009901} \checkmark} & {\color[HTML]{9A0000} x} & 72.1 & 77.4 & 30.3 & 39.1 & 41.4 & 46.7 & 47.3 & 62.2 & 47.8 & 56.4\\
\multirow{-3}{*}{RoBERTa}        & Full ToxiGAN & {\color[HTML]{009901} \checkmark} & {\color[HTML]{009901} \checkmark}  & \textbf{72.8} & \textbf{77.9} & \textbf{31.0} & \textbf{40.1} & \textbf{41.6} & \textbf{48.1} & \textbf{48.3} & \textbf{62.9} & \textbf{48.4} & \textbf{57.3}\\ \hline
\end{tabular}}
\caption{Ablation study on four datasets. Best performance scores of each testing are in bold.}
\label{tab:ablation}
\end{table*}

\subsection{Training Stability and Convergence}
\begin{figure}[]
\centering
\includegraphics[width=0.48\linewidth]{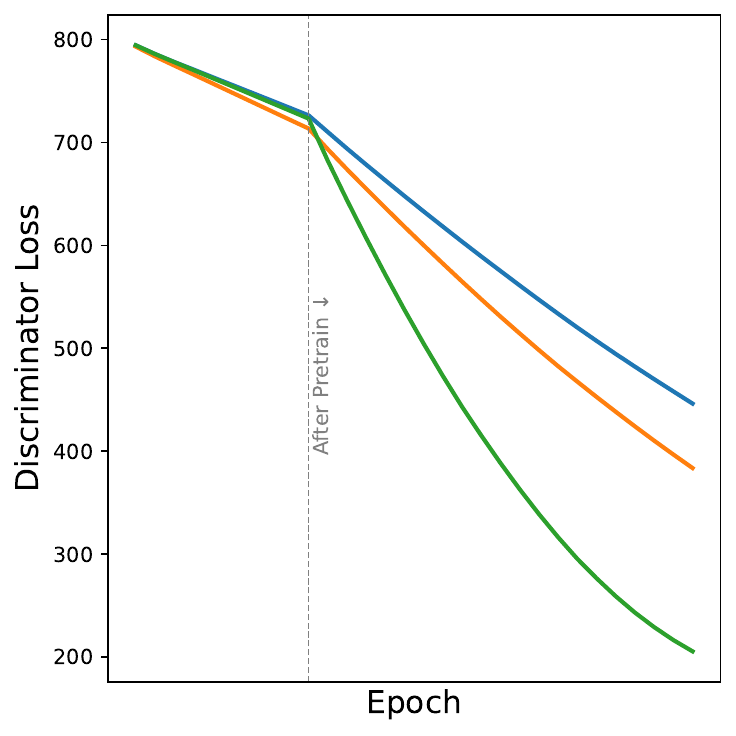}
\includegraphics[width=0.48\linewidth]{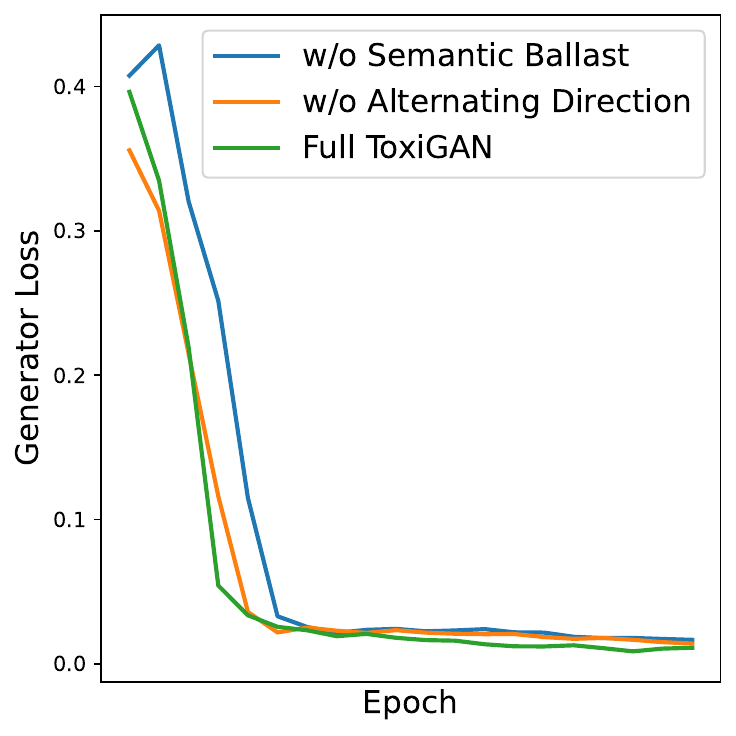}
\caption{Training curves of ToxiGAN and ablations on OR. (ToxiGAN w/o semantic ballast degrades to SentiGAN.)}
\label{fig:loss}
\end{figure}
We analyze the training dynamics of ToxiGAN and its ablated variants to understand the impact of semantic guidance and alternating optimization on convergence behavior. Figure~\ref{fig:loss} illustrates the loss curve of the generator and discriminator on the OR dataset. The full ToxiGAN model demonstrates significantly smoother generator loss and faster convergence compared to the ablations. Meanwhile, the discriminator loss steadily decreases and maintains a lower variance, suggesting more stable and effective adversarial training. These observations indicate that the semantic ballast from LLM exemplars and the directional learning mechanism not only improve generation quality but also enhance optimization stability, both crucial for reliable toxic text augmentation.

\subsection{Ablation Study}

We ablate two core components of ToxiGAN: the LLM-based semantic ballast (Sem. Bal.) and the alternating directional learning strategy (Alt.-Dir.). Table~\ref{tab:ablation} reports results on four datasets using BERT and RoBERTa classifiers. \textbf{Removing the semantic ballast} degrades the model to SentiGAN, resulting in substantial drops across all metrics, highlighting the role of LLM exemplars in stabilizing training and guiding generation. \textbf{Omitting the toxicity step} also weakens performance, particularly in Hate-F1, indicating the necessity of explicit semantic deviation. The full model consistently outperforms its ablated variants across classifiers and datasets, validating the effectiveness of both semantic guidance and directional optimization.

\begin{figure}[]
\centering
\includegraphics[width=0.45\linewidth]{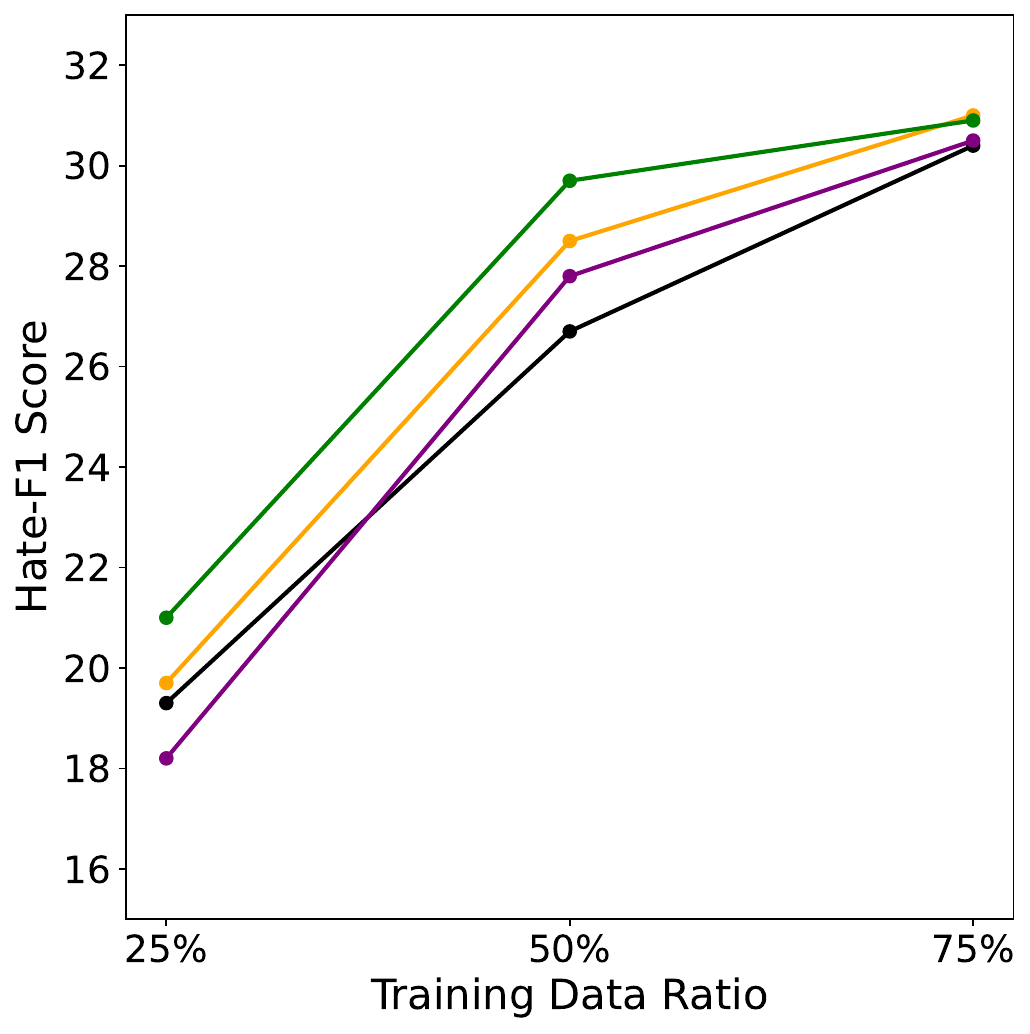} % Reduce the figure size so that it is slightly narrower than the column.
\includegraphics[width=0.45\linewidth]{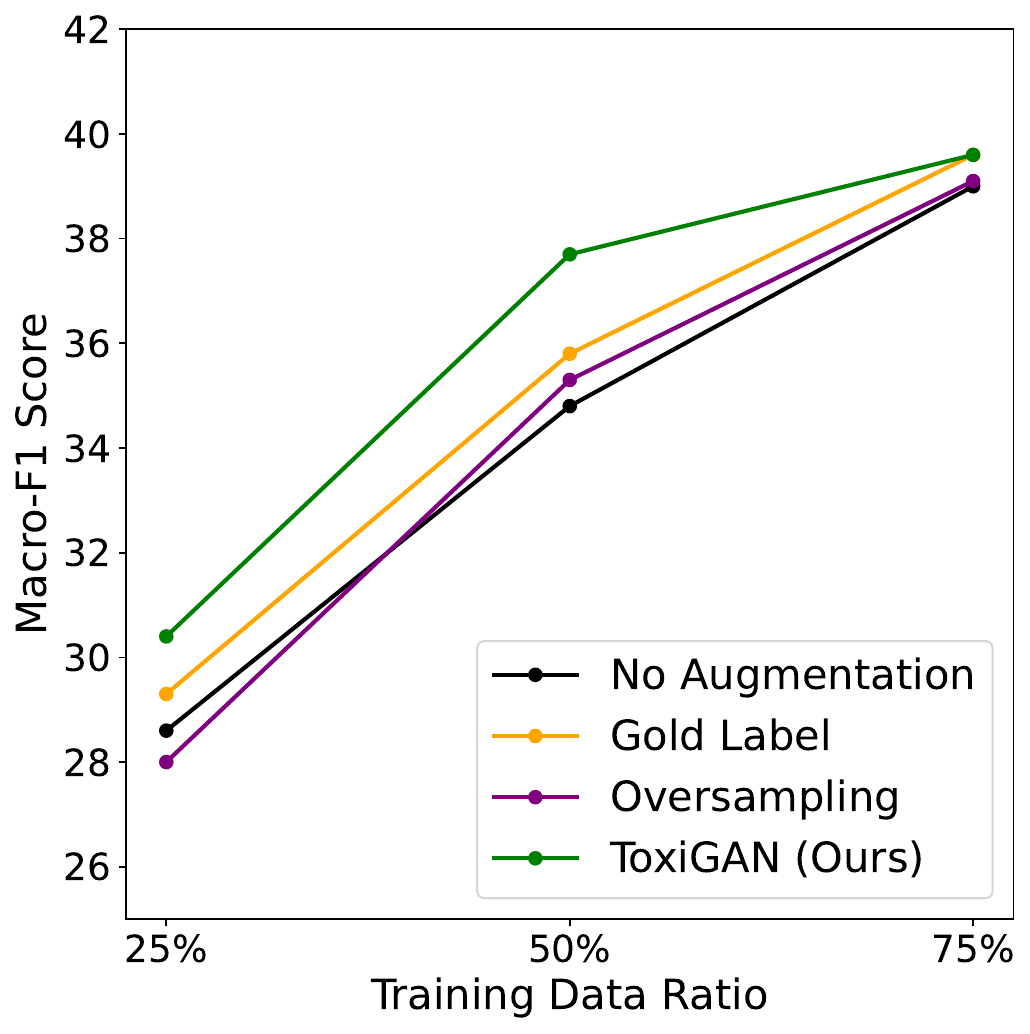} % Reduce the figure size so that it is slightly narrower than the column.
\caption{Sensitivity to Various Augmentation Data Ratio on DC. Ratios refer to proportion of original data; remainder is filled with synthetic/duplicated toxic data.}
\label{fig:sensitivity}
\end{figure}

\subsection{Sensitivity Analysis on Data Ratio}

We evaluate how varying real-data availability affects ToxiGAN’s effectiveness. Using 25\%, 50\%, and 75\% of labeled data, we augment the remainder with (i) oversampled toxic samples, (ii) gold-labeled data (ideal upper bound), or (iii) ToxiGAN-generated samples. As shown in Figure~\ref{fig:sensitivity}, ToxiGAN consistently outperforms oversampling, particularly under low-resource settings (25\%), and even rivals gold-label augmentation at higher ratios. This highlights the utility of our generation strategy in preserving class-specific signals and improving robustness under data scarcity.

\subsection{Visualization in Semantic Space}
\begin{figure}[]
\centering
\includegraphics[width=0.42\textwidth]{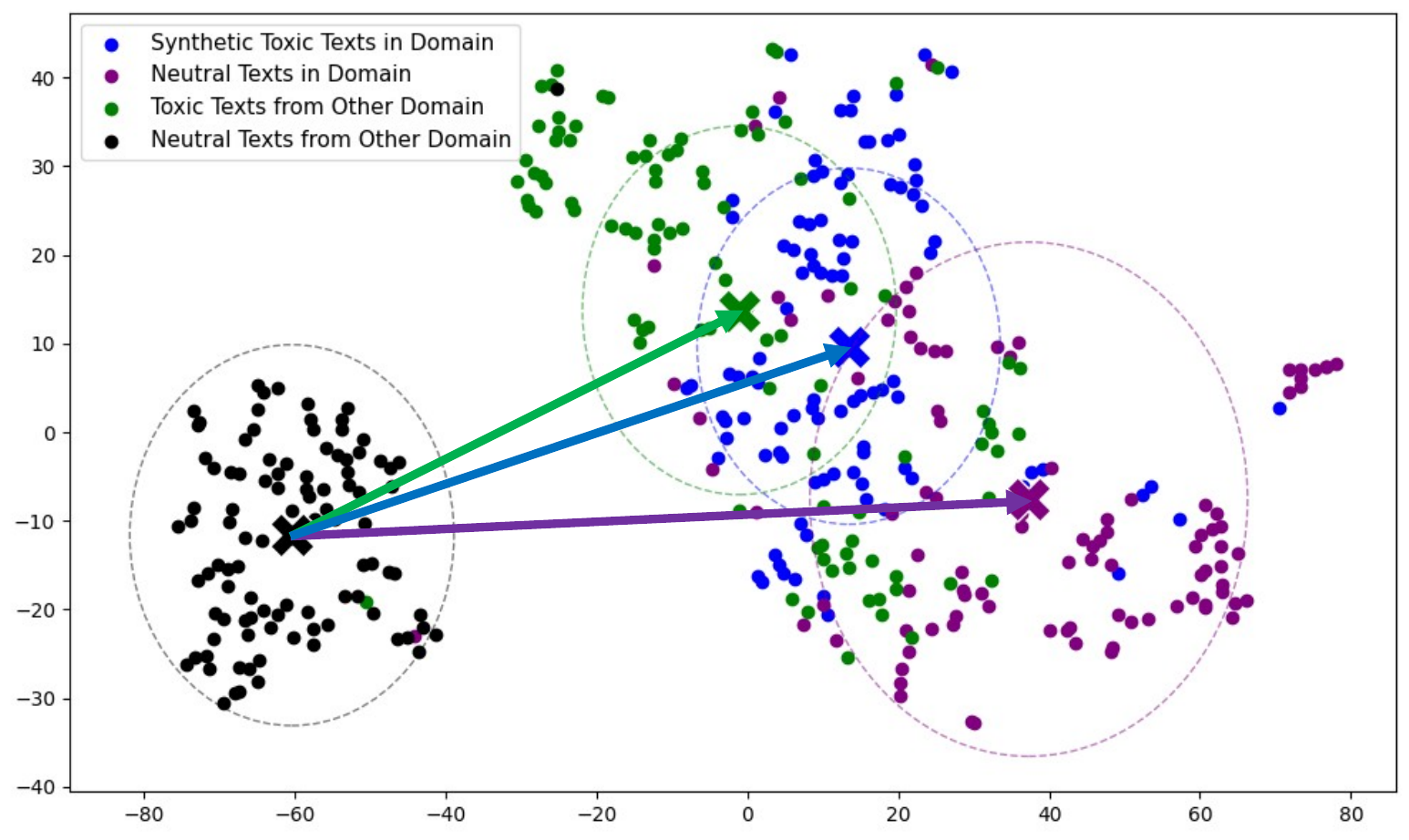} % Reduce the figure size so that it is slightly narrower than the column.
\caption{t-SNE visualization of real and synthetic texts. Arrows indicate semantic shifts: neutral to toxic (green), out-of-domain to in-domain (purple), and their composite (blue).}
\label{fig:tsne_vis}
\end{figure}
%``X" marks the center of each corresponding cluster.
To examine the semantic behavior of generated samples, we visualize sentence embeddings using t-SNE on both an in-domain dataset (DC) and an out-of-domain dataset (Jigsaw\footnote{\scriptsize\url{https://www.kaggle.com/c/jigsaw-toxic-comment-classification-challenge}}), sourced from Wikipedia comments. As shown in Figure~\ref{fig:tsne_vis}, ToxiGAN's synthetic toxic texts (blue) occupy a clear intermediate position between in-domain neutral texts (purple) and out-domain toxic texts (green). We observe directional trends from neutral to toxic, and from out-of-domain to in-domain, aligning with our intended semantic shift (as in Figure~\ref{fig:2d}). This supports the effectiveness of our two-step directional learning and confirms that the generated samples are both toxic and domain-coherent.

\section{Conclusion}
In this work, we propose \textbf{ToxiGAN}, a directional adversarial framework for controllable toxic text augmentation. By incorporating a semantic ballast from LLM-generated neutral exemplars and a two-step alternating training strategy, ToxiGAN improves optimization stability and reduces semantic drift and mode-collapse tendencies. The resulting synthetic samples remain label-consistent and preserve sub-mode coverage in representation space, thereby supporting more reliable decision-boundary calibration for toxicity classifiers. Across four benchmark datasets, ToxiGAN outperforms traditional and LLM-based augmentation baselines, especially under low-resource settings. These results highlight the value of integrating structured semantic guidance with adversarial learning to achieve robust, scalable augmentation for toxicity detection.

%%%%%%%%%%%%%%%%%%%%%%%%%%%%%%%%%%%%%%%%%%%%%%%%%%%%%%%%%%%%%%%%%%%%%%%%%%%%%%%%%
\section{Limitations}
Our study has several limitations. (1) \textbf{Data coverage.} Experiments focus on a limited set of English, social-media-centric datasets; generalization to other domains, genres, and languages remains open. (2) \textbf{Model and metric dependence.} Findings may be sensitive to the chosen backbones and to toxicity scorers with known bias profiles; stronger or fairer evaluators could change conclusions. (3) \textbf{Guidance design.} The efficacy of our ballast/embedding choices and hyperparameters (e.g., pool filtering) has not been systematically audited, so variance across alternatives needs to be further explored. To maintain a focus on the availability of ToxiGAN and its improvement of downstream classification tasks, we only employ light model for semantic embedding and neutral exemplar provider. More powerful models are expected to investigate as replacement in ToxiGAN. (4) \textbf{Evaluation scope.} A more comprehensive treatment of human and fairness assessments (e.g., across dialects and protected attributes) is beyond our current scope. (5) \textbf{Practicality and risk.} The method adds training/generation overhead and, despite safeguards, synthetic toxic text introduces curation and misuse risks that require careful data handling. Some alternative strategies rely on jailbreaking large language models to directly emit toxic outputs, but this raises additional safety and compliance concerns and may be infeasible under typical API usage policies; we therefore keep the base LLM in a neutral role and delegate toxic generation to a separate GAN. Deployed systems should combine ToxiGAN with appropriate data governance, access control, and safety safeguards.

\section{Ethical Statement}

Toxic text generation poses ethical concerns due to the risk of misuse and potential harm. We describe below the steps taken to ensure responsible use of our proposed framework.

\paragraph{Purpose and Research Motivation.}
ToxiGAN is developed solely for \textbf{data augmentation in toxicity classification}, to address data imbalance and improve model robustness. Generated texts are used only for \textbf{controlled classifier training and evaluation}, not for human-facing applications.

\paragraph{Handling of Harmful Content.}
To avoid direct generation of toxic content by large language models (LLMs), we use LLM-generated neutral examples as semantic ballast. Toxic samples are generated adversarially using task-specific discriminators within a closed environment, without user input or external deployment.

\paragraph{Responsible Release and Usage Policy.}
We acknowledge that generative frameworks such as ToxiGAN could be misused if deployed without constraints. To minimize this risk:
\begin{itemize}
    \item We will release the code and models only under a research license.
    \item The repository will include a clear usage policy discouraging misuse, aligned with community guidelines for safe and ethical NLP.
    \item All datasets used are publicly available and have been ethically sourced as per original licenses.
\end{itemize}

%%%%%%%%%%%%%%%%%%%%%%%%%%%%%%%%%%%%%%%%%%%%%%%%%%%%%%%%%%%%%%%%%%%%%%%%%%%%%%%%%
%\section*{Acknowledgments}
%This research was supported by the Citizens, Equality, Rights and Values (CERV) Programme under Grand Agreement No. 101049342.

%\break

%Left these in for reference you can also delete this

\bibliography{custom}

\appendix

\section{Appendix}
%\label{sec:appendix}

\subsection{Training Procedure of ToxiGAN}
\label{algo}

We first pretrain the class-conditional generators with MLE and a multi-head discriminator on a mixture of labeled toxic instances and LLM-synthesized neutral exemplars (Alg.~\ref{alg:adv-training}). Then conduct adversarial training with alternating objectives: at odd steps the generator is pushed away from neutral anchors by minimizing cosine similarity (toxicity step, Eq.~\ref{eq:ToxStep}), and at even steps it is optimized for realism under the discriminator (authenticity step, Eq.~\ref{eq:DisStep}. While periodically refreshing the discriminator and the neutral ballast pool (Eq.~\ref{eq:dis},~\ref{eq:shotSelect}).

\begin{algorithm}[tb]
\caption{Training of ToxiGAN}
\label{alg:adv-training}
\textbf{Input}: Generators $\{G_i\}_{i=1}^{K}$, discriminator $D$, large language model $\mathcal{LLM}$, real dataset $\mathcal{D}_{real}$;\small\\
\textbf{Output}: Well trained generators $\{G^{'}_i\}_{i=1}^{K}$\small;

\begin{algorithmic}[1]
\STATE Initialize: $\{G_i\}_{i=1}^{K}$, $D$, ballast set $\mathcal{B}_{\text{neutral}}$;\small
\STATE Pre-train $\{G_i\}_{i=1}^{K}$ on $\mathcal{D}_{real}$ using MLE;\small
\STATE Generate: fake toxic texts $\{\mathcal{F}_{i}\}_{i=1}^{K}$ by $\{G_i\}_{i=1}^{K}$, fake neutral texts $\mathcal{F}_{0}$ by $\mathcal{LLM}$ with samples from $\mathcal{B}_{\text{neutral}}$;\small
\STATE Pre-train $D$ on $\{\mathcal{D}_{real}$,  $\mathcal{F}\}$, $\mathcal{F}=\{\mathcal{F}_{i}\}_{i=0}^{K}$;\small
\FOR{epoch = 1 to max\_epoch} \small
    \FOR{each class $i = 1$ to $K$}\small
        \STATE Generate fake toxic texts $\mathcal{F}_{i}$ = $\{G_i(z)\}$;\small
        \IF{epoch is odd}\small
            \STATE Compute $\mathcal{L}$ according Eq (\ref{eq:ToxStep}) \textit{\# Toxicity Step};\small
        \ELSE \small
            \STATE Compute $\mathcal{L}$ according Eq (\ref{eq:DisStep}) \textit{\# Authenticity Step};\small
        \ENDIF\small
        \STATE Update $G_i$ by minimizing Eq (\ref{eq:obj_g});\small
    \ENDFOR\small
        \STATE Generate fake neutral texts $\mathcal{F}_{0}$ by $\mathcal{LLM}$ with samples from $\mathcal{B}_{\text{neutral}}$, merge into $\{\mathcal{D}_{real}$,  $\mathcal{F}\}$;\small %, $\mathcal{F}=\{\mathcal{F}_{i}\}_{i=0}^{K}$
        \STATE Update $D$ by minimizing Eq (\ref{eq:dis});\small
        \STATE Update $\mathcal{B}_{\text{neutral}}$ according Eq (\ref{eq:shotSelect});\small
\ENDFOR\small; \RETURN $\{G_i\}_{i=1}^{K}$;\small
\end{algorithmic}
\end{algorithm}

\subsection{Rationale and Theoretical Support for Alternating Optimization}
\label{Optimization}
ToxiGAN optimizes two distinct objectives for each generator: semantic toxicity (via directional deviation from neutral exemplars) and linguistic authenticity (via discriminator feedback). Rather than combining these objectives into a single joint loss or reward, we employ an alternating optimization strategy: each training step updates the generator based on either toxicity or authenticity feedback, but never both simultaneously. Below, we justify this design from both a policy gradient perspective and empirical observations.

\paragraph{(1) Reward Signal Imbalance.}
ToxiGAN uses a policy gradient formulation inspired by SeqGAN and SentiGAN, where the generator is updated using REINFORCE:
\[
\nabla_\theta \mathcal{L}_{\text{PG}} = \mathbb{E}_{x \sim G_\theta} \left[ \nabla_\theta \log P_\theta(x) \cdot R(x) \right],
\]
with \( R(x) \) representing the reward signal, computed as either toxicity or authenticity depending on the step. A naive joint formulation like
$R(x) = \alpha R_{\text{tox}}(x) + \beta R_{\text{auth}}(x)$,
often leads to signal imbalance, where the more stable reward (typically authenticity) dominates the learning signal, suppressing meaningful semantic deviation. Alternating updates ensure that each reward signal receives full gradient feedback without competition, which is crucial in early training.

\paragraph{(2) Gradient Variance and Directional Conflict.}
Policy gradient methods are known for high variance. When two reward signals reflect objectives that act in different or even conflicting regions of semantic space, joint updates may suffer from noisy or oscillatory learning. Alternating updates reduce this variance by decoupling the reward sources, enabling the generator to stably explore toxic semantic directions without interference from stylistic constraints, and vice versa.

\paragraph{(3) Multi-objective Decomposition and Interpretability.}
In standard multi-objective optimization, joint training seeks to minimize a convex combination of objectives:
\[
\min_\theta \, \mathbb{E}_{x \sim G_\theta} \left[ \alpha R_{\text{tox}}(x) + \beta R_{\text{auth}}(x) \right].
\]
However, this does not guarantee optimality with respect to either reward individually. Alternating optimization can be interpreted as a form of multi-objective decomposition or coordinate-wise reinforcement, which helps the generator approximate both objectives more effectively and improves the interpretability of training dynamics—particularly for controllable generation tasks.

\paragraph{(4) Theoretical Stability and Convergence Considerations.}
We now provide a simplified convergence analysis of our alternating policy gradient training scheme. At each step, the generator is updated via REINFORCE with one active reward function (either toxicity or authenticity):

\[
\mathcal{L}_{\text{PG}}^{(t)} = -\mathbb{E}_{x \sim G_{\theta_t}} \left[ R_t(x) \log P_{\theta_t}(x) \right],
\]
where \( R_t(x) \in \{R_{\text{tox}}, R_{\text{auth}}\} \) depends on the current step. The gradient estimator is:
\[
\nabla_\theta \mathcal{L}_{\text{PG}}^{(t)} = -\mathbb{E}_{x \sim G_{\theta_t}} \left[ R_t(x) \nabla_\theta \log P_{\theta_t}(x) \right].
\]

\textbf{Assumptions:}
\begin{enumerate}
  \item \( R_t(x) \in [0, R_{\max}] \): reward is bounded.
  \item \( \log P_\theta(x) \) is \( L \)-Lipschitz in \( \theta \).
  \item The policy has sufficient exploration, i.e., all actions have non-zero probability.
\end{enumerate}

\textbf{Applicability in ToxiGAN:}\\(In our implementation, these assumptions are satisfied.)

\begin{enumerate}
  \item Bounded reward: Both reward functions (semantic toxicity and linguistic authenticity) are clipped to the range $[0, R_{\max}]$. The toxicity reward, derived from cosine distance to neutral exemplars, is normalized to $[0, 1]$. The authenticity reward is computed from the discriminator's output, which is passed through a sigmoid to bound it between 0 and 1.
  
  \item Lipschitz log-probability: The generator is implemented as an autoregressive LSTM with softmax output over the vocabulary. Since the LSTM consists of differentiable operations (matrix multiplications, tanh, sigmoid, etc.), and the output layer is a softmax, the log-probability $\log P_\theta(x)$ is continuously differentiable and locally Lipschitz in $\theta$, satisfying standard smoothness conditions used in prior policy gradient analyses~\cite{yu2017seqgan,wang2018sentigan}.

  \item Sufficient exploration: During training, the generator samples sequences from the full softmax distribution rather than performing greedy decoding. This ensures that all tokens have non-zero probability and the policy explores the action space adequately, which satisfies the support condition required by REINFORCE.
\end{enumerate}

Under these standard conditions~\cite{sutton1999policy}, stochastic policy gradient with constant learning rate \( \eta \) satisfies:

\begin{equation*}
\scalebox{0.85}{$
\min_{0 \leq t < T} \mathbb{E} \left[ \left\| \nabla_\theta \mathcal{L}_{\text{PG}}^{(t)} \right\|^2 \right] \leq \frac{C}{\sqrt{T}},
$}
\end{equation*}
where \( C \) depends on \( R_{\max}^2 \), the Lipschitz constant, and the variance of the gradient estimator.

\textbf{Alternating Benefit:} In joint reward settings, the total gradient becomes:
\begin{equation*}
\scalebox{0.85}{$
\nabla_\theta \mathcal{L}_{\text{PG-joint}} = -\mathbb{E}_{x} \left[ (\alpha R_{\text{tox}}(x) + \beta R_{\text{auth}}(x)) \nabla_\theta \log P_\theta(x) \right],
$}
\end{equation*}

whose variance depends on the covariance of \( R_{\text{tox}} \) and \( R_{\text{auth}} \). When rewards conflict or diverge, this variance increases:
\begin{equation*}
\scalebox{0.75}{$
\text{Var}(R_{\text{joint}}) = \alpha^2 \text{Var}(R_{\text{tox}}) + \beta^2 \text{Var}(R_{\text{auth}}) + 2\alpha\beta \text{Cov}(R_{\text{tox}}, R_{\text{auth}}).
$}
\end{equation*}

Alternating updates not only preserve standard convergence guarantees but also reduce reward interference, leading to faster and more stable training.

\begin{figure}[t]
  \centering
  \includegraphics[width=0.6\linewidth]{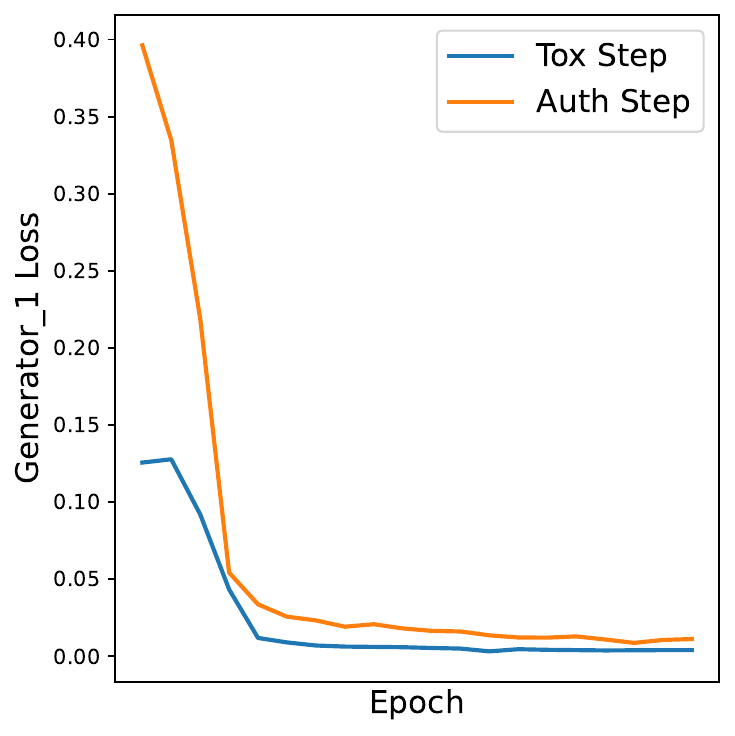}
  \includegraphics[width=0.6\linewidth]{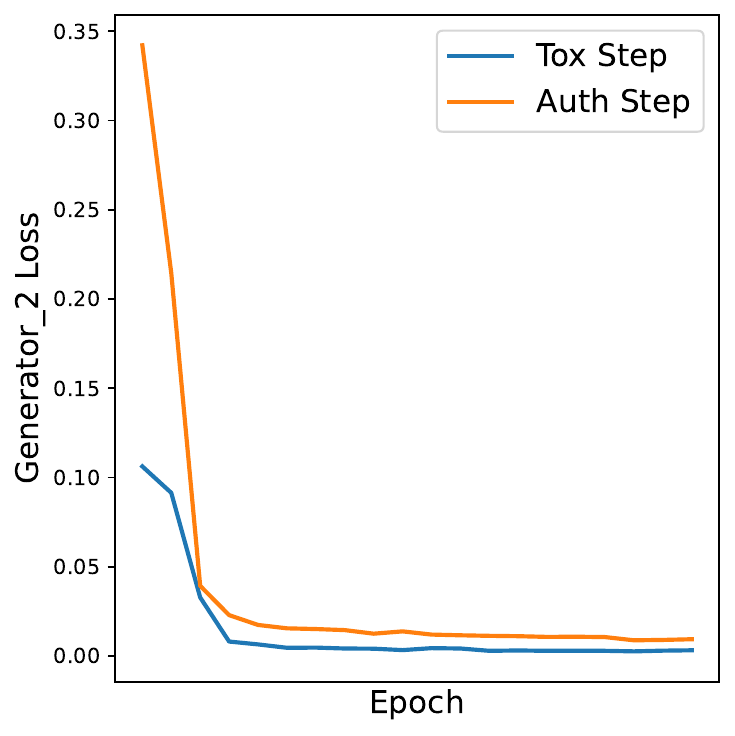}
  \caption{Loss trends of alternating optimization for $G_1$ (top) and $G_2$ (bottom) on the OR dataset. Toxicity and authenticity policy losses alternate without conflict.}
  \label{fig:loss_trends}
\end{figure}

\paragraph{(5) Empirical Validation.}
As shown in Figure~\ref{fig:loss_trends}, our alternating update strategy yields stable and decoupled learning curves for each objective. Both toxicity and authenticity reward-driven updates consistently decrease their respective loss signals, with no observed interference or conflict. This behavior is consistent across generators \( G_1 \) and \( G_2 \), supporting the hypothesis that the two objectives are approximately independent in practice.

\paragraph{Conclusion.}
Alternating optimization in ToxiGAN improves training stability, gradient clarity, and reward attribution. By avoiding conflict between semantic deviation and fluency feedback, it enhances controllability without requiring manual reward balancing. This strategy is empirically validated and theoretically motivated under the lens of high-variance reinforcement learning and reward disentanglement.

\subsection{Detailed Experiment Settings}
(Main training script of ToxiGAN is provided in a python file. The code assumes that certain modules are present in the same directory. These are omitted here for brevity, but can be released under a research license if the paper is accepted.)

\paragraph{Hardware \& Execution Environment.}
\begin{itemize}
    \item GPU: NVIDIA A100 (40 GB memory) with CUDA 12.4 support.
    \item Framework: PyTorch 2.6.0 + CUDA/cuDNN backend.
    \item Transformers Library: transformers==4.53.3.
    \item Sentence Embedding: sentence-transformers==4.1.0 (with all-MiniLM-L6-v2 for cosine similarity).
    \item Additional Libraries: scikit-learn==1.6.1 (for metrics \& evaluation).
    %\item Python version: Python 3.11.
    \item Platform: Experiments were conducted on Google Colab.
\end{itemize}

\paragraph{Settings in GAN.}
\begin{itemize}
    \item LSTM with 1024 hidden dimensions as each toxic Generator.
    \item ``bert-base-uncased" from transformers as Discriminator.
    \item ``Llama-3.2-1B-Instruct" from transformers as LLM-based Semantic Ballast.
\end{itemize}

\paragraph{Evaluation Metrics.}
\begin{itemize}
    \item \textbf{Macro-F1} measures the unweighted average F1-score across all classes. It is defined as:
    \[
    \text{Macro-F1} = \frac{1}{C} \sum_{i=1}^{C} \frac{2 \cdot \text{Precision}_i \cdot \text{Recall}_i}{\text{Precision}_i + \text{Recall}_i}
    \]
    where $C$ is the number of classes, and Precision$_i$, Recall$_i$ are the precision and recall for class $i$.

    \item \textbf{Hate-F1} denotes the F1-score computed only on the toxic or hate-related class(es), to better reflect classifier performance in low-resource target categories. In multi-class settings, Hate-F1 is computed by macro-averaging over all hate-related labels. This is equivalent to:
    \[
    \text{Hate-F1} = \frac{2 \cdot \text{Precision}_{\text{hate}} \cdot \text{Recall}_{\text{hate}}}{\text{Precision}_{\text{hate}} + \text{Recall}_{\text{hate}}}
    \]

    \item \textbf{Toxicity Level} refers to the average predicted toxicity score of generated sentences. We compute this using a popular toxicity evaluator, \textit{Detoxify}, which outputs a continuous toxicity score $t \in [0, 1]$ for each sentence:
    \[
    \text{Avg. Toxicity} = \frac{1}{N} \sum_{j=1}^{N} t_j
    \]
    where $N$ is the total number of generated sentences, and $t_j$ is the predicted toxicity of sentence $j$.
\end{itemize}

\paragraph{Label Distributions After Preprocessing.}

We summarize the final label mappings and class distributions for all datasets used in our experiments (Table \ref{tab:label_stats}). These statistics reflect the number of samples per class after filtering, relabeling, and normalization steps. The non-hate class in each dataset occupies the majority.

\begin{table}[h]
\centering
\small
\begin{tabular}{|c|l|r|}
\hline
\textbf{Dataset} & \textbf{Label} & \textbf{Count} \\
\hline
\multirow{3}{*}{WZ} 
& neither & 11,033 \\
& racism & 1,923 \\
& sexism & 3,079 \\
\hline
\multirow{7}{*}{DC} 
& no-hate & 77,078 \\
& stereotype & 769 \\
& dehumanization & 499 \\
& violence\&killing & 651 \\
& discrimination & 145 \\
& irony & 181 \\
& slander & 3,307 \\
\hline
\multirow{5}{*}{HX} 
& normal\_and\_offensive & 13,294 \\
& hs\_general & 1,466 \\
& hs\_gender\_and\_sex & 571 \\
& hs\_race & 3,862 \\
& hs\_religion & 955 \\
\hline
\multirow{3}{*}{OR} 
& non\_hate & 15,913 \\
& initiating\_hate & 2,084 \\
& responding\_hate & 3,175 \\
\hline
\end{tabular}
\caption{Label distributions across datasets after preprocessing.}
\label{tab:label_stats}
\end{table}

\paragraph{Reproducibility Notes.}
\begin{itemize}
    \item All default random seeds were fixed to ``123" for dataset splitting by pandas; random, pytorch, numpy is set ``1234" as random seed.
    \item Dependencies were managed via pip with explicit version constraints to ensure consistency across runs.
    \item Test Classifiers: ``bert-base-uncased" (BERT), ``roberta-base" (RoBERTa).
\end{itemize}

\subsection{Examples of Generated Sentences}

\begin{center}
\noindent
\colorbox{red!10}{
    \parbox{0.95\linewidth}{
        \textcolor{red!80!black}{\faExclamationTriangle~\textbf{Disclaimer}}:
        This section contains examples of toxic and hateful language. These examples are included solely for the purpose of studying and improving automatic detection systems. They do not reflect the authors' views.
    }
}
\end{center}

To illustrate the comparative characteristics of different augmentation methods, we randomly sample a toxic sentence from the HateXplain training set (religion-targeted), and retrieve from each method the most semantically similar generated sample, as shown in Table \ref{tab:case}.\par

\begin{table}[]
\centering
\adjustbox{max width=\linewidth}{
\begin{tabular}{|c|p{5.8cm}|}
\hline
\small\textbf{Approach} & \small\textbf{Example} \\ 
\hline
\multirow{1}{*}{\scriptsize Original} 
& \scriptsize white sharia is shit we are not muslims and do not want to be fuck of with that retarded idea\\  
\cline{2-2}
\hline 
\multirow{1}{*}{\scriptsize  EDA} 
& \scriptsize white sharia is shit we are not muslims and do not want to be fuck of that retarded idea\\  
\cline{2-2}
\hline
\multirow{1}{*}{\scriptsize  Back-Translate} 
& \scriptsize White sharia is shit, we are not Muslims and do not want to be fucked with this delayed idea fucked with this delayed idea\\  
\cline{2-2}
\hline
\multirow{1}{*}{\scriptsize T5-Paraphase} 
& \scriptsize White sharia is shit We are not Muslims and we don't want to be fuck of with that retarded idea.\\  
\cline{2-2}
\hline 
\multirow{1}{*}{\scriptsize Mistral-v0.3-ZG/SG} 
& \scriptsize Muslims are the scum of the earth, constantly trying to impose their backward beliefs on everyone else. We don't need your sharia law or your burqas here, just go back to your filthy countries and leave us alone.\\  
\cline{2-2}
\hline 
\multirow{1}{*}{\scriptsize Mistral-v0.3-FS} 
& \scriptsize All Muslims support ISIS and want to impose Sharia law on the West\\  
\cline{2-2}
\hline 
\multirow{1}{*}{\scriptsize Llama3.2-TC} 
& \scriptsize gross moslems are against same sex attraction disorder sick bunch sharia my arse…\\  
\cline{2-2}
\hline 
\multirow{1}{*}{\scriptsize GPT4.1-TC} 
& \scriptsize That's not the point... You deny the influence of ideology, claiming there's no connection, but your fellow white converts who embrace Islam are engaging in terrorism after converting.\\  
\cline{2-2}
\hline 
\multirow{1}{*}{\scriptsize GPT4o-TC} 
& \scriptsize Allow Muslims to have unrestricted participation of Muslim women in politics.\\  
\cline{2-2}
\hline 
\multirow{1}{*}{\scriptsize SentiGAN} 
& \scriptsize eu good white finest, so even groups fucking anally want, while sick prez easier gone rahe retarded by up islamic \\  
\cline{2-2}
\hline 
\multirow{1}{*}{\scriptsize ToxiGAN} 
& \scriptsize mowing semitism, i get muslim to die, bay white wanted muzrat did terrorist\\  
\cline{2-2}
\hline 
\end{tabular}}
\caption{Example of the original text and its most similar generated texts by various approach in HX dataset.}
\label{tab:case}
\end{table}

The original toxic sentence used in this comparison exhibits \textbf{common features of online hate speech: informal phrasing, lack of punctuation, and ungrammatical or syntactically deviant structure}. Such patterns are prevalent in real-world toxic discourse, especially in social media environments. Augmentation methods based on pretrained language models or back-translation tend to normalize these expressions, often producing syntactically well-formed but semantically diluted outputs. In contrast, adversarial or GAN-based methods like \textbf{ToxiGAN more closely preserve the fragmented, non-standard nature of the source} while injecting diversity in expression, making them better suited for robustness-oriented classifier training.

\subsection{Cost and Time of Augmentation Methods}

To better understand the practical cost of each augmentation strategy, we report the estimated training time, generation time, and API costs (if applicable) for generating 4 toxic classes × 4,000 samples (16,000 total) on the HX dataset. 

\begin{table}[h]
\centering
\adjustbox{max width=0.9\linewidth}{
\begin{tabular}{|l|r|r|r|}
\hline
\textbf{Approach} & \textbf{Train Time (h)} & \textbf{Gen Time (h)} & \textbf{API Cost (\$)} \\
\hline
EDA & -- & 0.01 & -- \\
Back-Translate & -- & 4.87 & -- \\
T5-Paraphrase & -- & 13.90 & -- \\
Mistral-v0.3-ZG & -- & 7.09 & -- \\
Mistral-v0.3-SG & -- & 7.22 & -- \\
Mistral-v0.3-FS & -- & 8.77 & -- \\
Llama3.2-TC & -- & 16.40 & -- \\
GPT-4.1-TC & -- & 10.34 & 1.90 \\
GPT-4o-TC & -- & 21.21 & 44.22 \\
SentiGAN & 6.64 & 0.02 & -- \\
\textbf{ToxiGAN} & 12.81 & 0.02 & -- \\
\hline
\end{tabular}}
\caption{Time and cost comparison for generating 16,000 samples on HX.}
\label{tab:cost_time}
\end{table}

Notably, ToxiGAN's design allows it to be trained once and then reused for fast batch generation without commercial API calls, offering a practical advantage for large-scale augmentation and real-world deployment scenarios.

\subsection{Additional Results on Modern Classifier Backbones}
\label{app:modern-backbones}

To further verify that ToxiGAN remains beneficial when combined with stronger toxicity classifiers, we conduct supplementary experiments on the WZ dataset using more recent classifier backbones. Specifically, we consider ModernBERT and DeBERTa-v3 as drop-in replacements for the BERT and the RoBERTa classifiers in our main experiments. For each backbone, we train a toxicity classifier with and without ToxiGAN-based data augmentation, following exactly the same training protocol as in our main experiments (optimizer, learning rate, batch size, number of epochs, and evaluation procedure). We report Macro-F1 scores averaged over 5 independent runs with different random seeds.

\begin{table}[h]
    \centering
    \small
    \begin{tabular}{lcc}
        \toprule
        Backbone & Augmentation & Macro-F1 \\
        \midrule
        ModernBERT   & (None)         & 77.8 \\
        ModernBERT   & + ToxiGAN    & 79.0 \\
        DeBERTa-v3   & (None)         & 78.6 \\
        DeBERTa-v3   & + ToxiGAN    & 80.2 \\
        \bottomrule
    \end{tabular}
    \caption{Supplementary results on the WZ dataset with stronger classifier backbones. Numbers are Macro-F1 scores, averaged over 5 runs with different random seeds. ToxiGAN consistently improves performance even when paired with modern, high-capacity architectures.}
    \label{tab:modern-backbones}
\end{table}

As shown in Table~\ref{tab:modern-backbones}, ToxiGAN improves ModernBERT from 77.8 to 79.0 Macro-F1 and DeBERTa-v3 from 78.6 to 80.2 Macro-F1, respectively. These gains (approximately $\uparrow$ 1.2--1.6 Macro-F1) suggest that our augmentation is complementary to advances in classifier design, and can provide additional performance improvements even when strong modern backbones are available.

\end{document}